\documentclass[12pt]{article}
\usepackage{authblk}

\usepackage[a4paper,top=4cm,bottom=2cm,left=3cm,right=3cm,marginparwidth=1.75cm]{geometry}
\usepackage{setspace}
\usepackage{amsmath}
\usepackage{graphicx}
\usepackage[colorlinks=true, allcolors=blue]{hyperref}
\usepackage{natbib}
\usepackage{float}
\usepackage{svg}

\title{LLMs for clinical risk prediction}
\author[1]{Mohamed Rezk}
\author[1]{Patricia Cabanillas Silva}
\author[1]{Fried-Michael Dahlweid}
\affil[1]{Dedalus Healthcare, Antwerp, Belgium}

\begin{document}
\maketitle

\renewcommand\abstractname{Abstract}

\begin{abstract}
This study compares the efficacy of GPT-4 and clinalytix Medical AI in predicting the clinical risk of delirium development. Findings indicate that GPT-4 exhibited significant deficiencies in identifying positive cases and struggled to provide reliable probability estimates for delirium risk, while clinalytix Medical AI demonstrated superior accuracy. A thorough analysis of the large language model's (LLM) outputs elucidated potential causes for these discrepancies, consistent with limitations reported in extant literature. These results underscore the challenges LLMs face in accurately diagnosing conditions and interpreting complex clinical data. While LLMs hold substantial potential in healthcare, they are currently unsuitable for independent clinical decision-making. Instead, they should be employed in assistive roles, complementing clinical expertise. Continued human oversight remains essential to ensure optimal outcomes for both patients and healthcare providers.
\end{abstract}

\section*{Introduction}

 The biomedical field's burgeoning interest in natural language processing (NLP) is unsurprising, given its transformative potential for healthcare. Clinical language models offer significant promise, as they can be pre-trained on a wide range of biomedical resources, including scientific literature, real-world clinical data, or a combination thereof [\cite{omiye_large_2024}].

A notable example is Med-PaLM [\cite{singhal2023large}], the first AI system to surpass the passing score on the United States Medical Licensing Examination (USMLE). This achievement was further advanced with the release of Med-PaLM 2 in 2023, which scored an impressive 86.5\% on USMLE-style questions [\cite{singhal2023towards}]. Another significant development is Med-PaLM M, a multimodal version capable of processing and generating information from various medical data sources, including images, electronic health records (EHRs), and genomics data [\cite{tu_towards_2023}]. These advancements underscore the growing impact on the biomedical field.

These rapid advancements prompt the question of whether large language models (LLMs) can be effectively utilized in clinical settings for risk predictions and evaluations. While current models caution against their use in clinical practice, the examples and demonstrations presented are designed to build confidence in these technologies. The goal is to reach a point where we can reliably depend on such models, with the potential for them to eventually match or even exceed the performance of clinicians.

This study aims to compare the performance of LLMs to that of clinalytix Medical AI in performing clinical risk predictions. We employed GPT-4 as a representative LLM and focused on delirium as an exemplar use case. We compared the performance of the two systems and further investigated the LLM's output to gain a better understanding of the generated content and to determine the level of confidence that can be placed in such models.

\section*{Methodology}
We analyzed data from 190 cases, encompassing both positive and negative delirium patients. This dataset included unstructured raw text from electronic health records (EHRs) alongside structured data such as laboratory results, medication records, and vital signs. For clinalytix Medical AI, we utilized the standard calibration pipeline (Medical AI 4.0). Regarding the large language model (LLM), we converted all data into a raw text format compatible with the model. In cases where the text data exceeded the LLM's context window, we reduced the input by removing the oldest data until it fit within the context window's size constraints. The LLM was then tasked with assessing each case for the risk of developing delirium, providing a probability score for the likelihood of delirium development, and offering explanations for its positive or negative predictions.

\section*{Results}

\begin{figure}[H]
\centering
\includegraphics[width=1\linewidth]{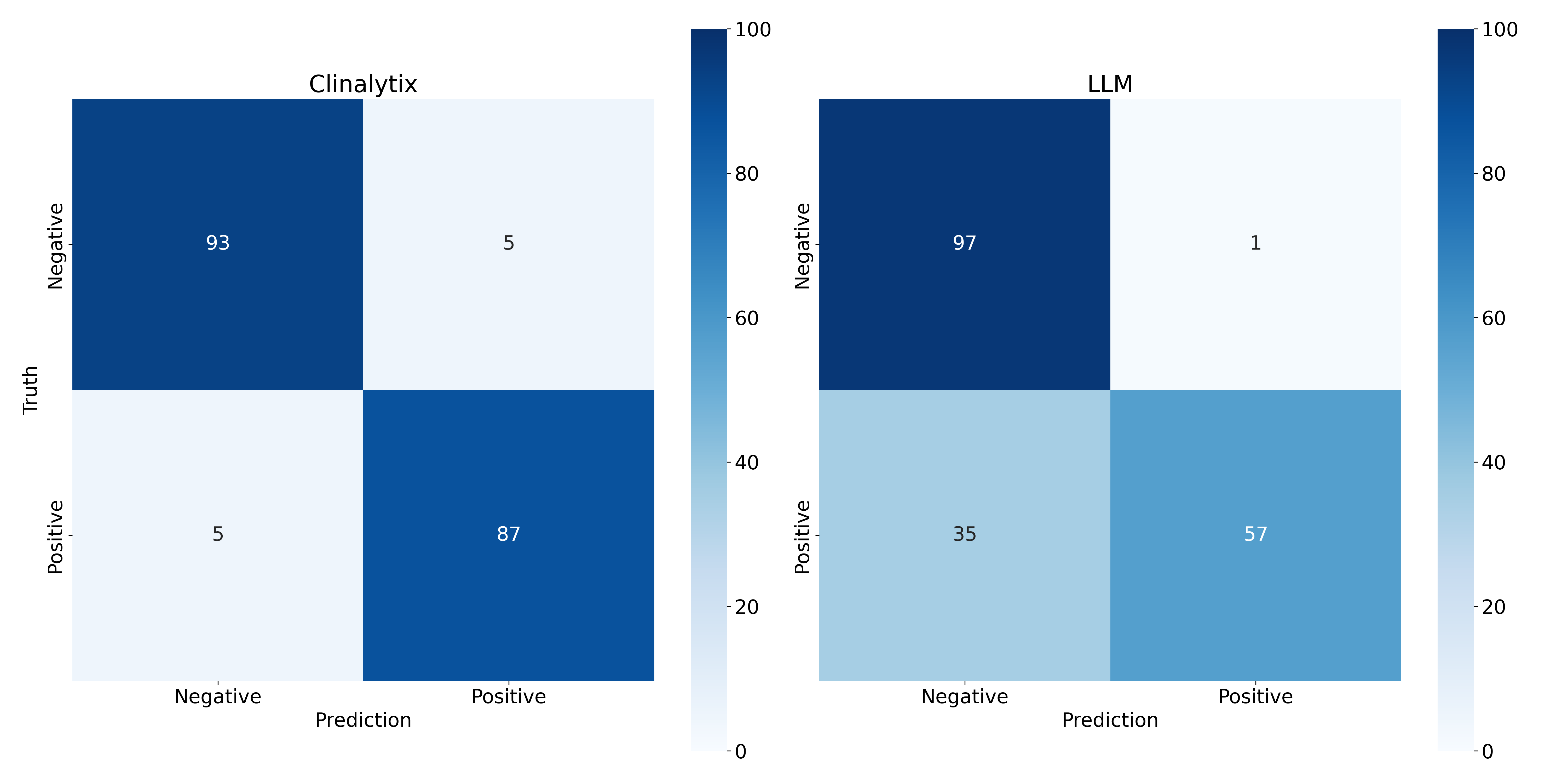}
\caption{\label{fig:confusion_matrix}Confusion Matrix Comparison of GPT-4 and clinalytix Medical AI in Predicting Delirium Risk. The figure illustrates the confusion matrices for both models, showing the distribution of true positives, true negatives, false positives, and false negatives. The comparison highlights the differences in the models' ability to accurately predict the clinical risk of delirium.}
\end{figure}

The comparison of clinalytix Medical AI and the LLM revealed a significant disparity in performance, with clinalytix Medical AI demonstrating markedly higher accuracy in predicting the risk of developing delirium (Table \ref{tab:metrics}). The LLM produced only one false positive, incorrectly predicting one case as at risk for delirium. However, it failed to identify approximately 38\% of the true positive cases, predicting them as negative, which resulted in a high number of false negatives (Figure \ref{fig:confusion_matrix}). Overall, the LLM exhibited very low recall, both in absolute terms and relative to clinalytix Medical AI.

\begin{table}[H]
\centering
\begin{tabular}{l|c|c}
Metric & clinalytix & LLM \\\hline
precision & 94.57\% & 98.28\%\\
recall  &  94.57\% & 61.96\% \\
f1-score & 94.57\% & 76.00\% \\
specificity & 94.90\% & 98.98\%
\end{tabular}
\caption{\label{tab:metrics}Evaluation Metrics for GPT-4 and clinalytix Medical AI models in Predicting Delirium Risk. This table compares the performance of the two models, highlighting their precision, recall, specificity, and F1-scores in identifying patients at risk of developing delirium.}
\end{table}

Predicting delirium risk is only the initial step; additional information is crucial to determine the appropriate course of action. This includes understanding the probability of delirium development and the rationale behind a model's prediction. clinalytix Medical AI goes beyond mere predictions by providing a calibrated probability of delirium risk. Model calibration ensures that the probability output accurately reflects the real-world likelihood of delirium occurrence. Conversely, LLMs do not provide a calibrated probability for delirium risk. Even when explicitly prompted, the model often responded that insufficient information was available to provide a reliable numerical probability.

Explainability is a crucial aspect of clinical risk prediction models, as it helps users understand the basis for a model's predictions. Various methods have been developed to elucidate why a model produces certain outcomes. Among these, LIME (Local Interpretable Model-agnostic Explanations) and SHAP (SHapley Additive exPlanations) are widely used techniques for interpreting complex machine learning models. Both methods have been employed in different versions of clinalytix Medical AI to improve transparency and build trust in the model's predictions. LIME simplifies the model's behavior locally [\cite{lime}], while SHAP provides consistent feature importance values [\cite{shap}]. For large language models (LLMs), explanations for predictions are generated as requested. However, these explanations should be interpreted with caution. The LLM's generative nature means that while it may produce plausible explanations, there is no guarantee that these reflect the true rationale behind its predictions or that there are no hallucinations [\cite{ji2023survey}]. This can lead to the generation of reasonable-sounding, yet potentially misleading explanations, making it essential to critically assess their validity.

Further investigation of the incorrectly predicted cases suggests some possible explanations:

\begin{itemize}
\item Context window limitations: The LLM has a fixed context window (8,000 tokens for GPT-4). The data size of a single patient can easily exceed that limit, and forcing the LLM to consider only the most recent data that fits within the context window can lead to loss of important information, resulting in erroneous predictions.

\item Risk prediction vs. detection: The model sometimes provided correct explanations and identified relevant terms but avoided indicating a risk of delirium if no delirium had already occurred. In such situations, the LLM acted more as a delirium detector rather than a risk predictor, despite the prompts clearly specifying risk prediction.

\item Data type preference: The model appeared to rely more heavily on textual information (raw text, medication) but seemed less adept at interpreting laboratory results, which is a known challenge for LLMs in making interpretations from raw values. Complex reasoning and inference from these results are not reliably performed by the LLM.

\end{itemize}

It is important to note that given the complex nature of LLMs, it is challenging to definitively pinpoint the reasons behind specific predictions.

\section*{Discussion}

Our investigation revealed that relying solely on an LLM to assess the clinical risk of delirium is not yet reliable, indicating that LLMs are currently ill-equipped for such complex tasks. Using GPT-4 as a representative model, we observed that it tended to avoid indicating delirium risk unless there were very strong indicators, resulting in a significant number of missed positive cases. Through manual analysis of the data, we identified several potential reasons for these shortcomings, including the limitations of the context window, inherent model behaviors, and the tendency to focus on specific pieces of information while overlooking others. These findings underscore the need for caution when applying LLMs to critical clinical evaluations.

\subsection*{Context window}

During our investigation, we utilized GPT-4, which had a context window of 8,000 tokens. Since the data for a single case or admission could exceed this limit, it was necessary to truncate the data to fit within the model's context window. This truncation could potentially result in the loss of critical information, which might be valuable for the LLM to accurately assess a case.

Interestingly, rapid advancements in LLMs have led to the release of models with significantly larger context windows. For instance, Anthropic's Claude 2.1 offers a context window of 200,000 tokens. While this might seem like a solution to the problem, it isn't necessarily the case. Research indicates that when dealing with large context windows, LLMs tend to focus on information presented at the beginning or end of the input, often neglecting details in the middle. This led to degraded performance when LLMs needed to access relevant information in the middle of long contexts [\cite{liu_lost_nodate}]. Additionally, larger context windows introduce increased computational complexity, as the complexity scales quadratically with the size of the context window. These factors suggest that simply expanding the context window may not fully address the challenges of processing large datasets in LLMs.

\subsection*{Model performance}

A critical question is whether these results are specific to a particular model or if different models might yield different outcomes. A recent study published in Nature Medicine demonstrated that even the most advanced LLMs struggle to accurately diagnose patients across various pathologies and consistently perform significantly worse than physicians. Additionally, LLMs do not adhere to diagnostic or treatment guidelines, leading the study to conclude that LLMs are not yet ready for autonomous clinical decision-making [\cite{hager_evaluation_2024}].

This raises an important question: why do these models perform so poorly compared to clinicians, despite recent demonstrations showing LLMs excelling in medical question-answering tasks, even achieving high scores on the USMLE exams? The answer lies in the fundamentally different nature of these scenarios. Exam questions or simulated problems are typically narrow in scope, focusing on specific areas or domains, making it easier for an LLM to analyze the limited data and arrive at a correct answer. In contrast, real-world clinical data is vast and diverse, encompassing a wide range of sources such as EHRs, lab results, and vital signs. This complexity and diversity likely impact the LLM's performance. Moreover, it has been shown that LLM performance can be influenced by the placement of relevant information within the text. While more data is generally advantageous in clinical practice, the effectiveness of LLMs may diminish if large portions of the data are not useful or if the model does not focus on crucial information due to its position within the input [\cite{liu_lost_nodate}]. Surprisingly, providing the LLM with more information actually deteriorated its diagnostic performance [\cite{hager_evaluation_2024}]. This presents a significant challenge in clinical settings, where physicians gather extensive information to comprehensively understand a patient's health status. The ability to discern the most salient details from this wealth of data is crucial, and the LLM's difficulty in doing so raises concerns about its efficacy in real medical situations. Therefore, the very nature of real-world data, with its complexity, may present challenges that LLMs are not yet equipped to handle effectively.

\subsection*{Model behaviour}

The behavior of LLMs can vary significantly for several reasons. As previously mentioned, the model's performance can be influenced by the location of relevant information within the input [\cite{liu_lost_nodate}]. Remarkably, presenting the exact same information in a different order can lead to substantial differences in the LLM's performance [\cite{hager_evaluation_2024}]. This inconsistency is unlikely to occur with clinicians, highlighting potential issues with LLMs and the need for caution when interpreting their results.

Moreover, our analysis revealed that the LLM tended to prioritize textual data over tabular information. This is particularly noteworthy because recent studies suggest that LLMs struggle to independently interpret laboratory results [\cite{hager_evaluation_2024}]. The complexity may increase if there are interactions between different pieces of information that are not well comprehended by the model. These factors highlight the current limitations of LLMs in clinical settings and the need for careful consideration before relying on their outputs.

More concerning is the generative nature of LLMs, which allows them to produce text that sounds coherent and reasonable, but with no guarantee of factual accuracy. Recent studies have identified significant discrepancies between the differential diagnoses made by LLMs and those made by radiologists [\cite{nakaura2024preliminary}]. A systematic review of ChatGPT's performance in the radiology field [\cite{keshavarz2024chatgpt}] found that while LLMs could provide appropriate recommendations, they rarely included nuanced or inferred guidance. These limitations have led researchers to conclude that human oversight remains crucial when using LLMs in clinical practice [\cite{woo2024evaluation}]. This issue extends beyond unimodal LLMs to multimodal models like GPT-4V, which process both text and images. These multimodal LLMs also face significant challenges in accurately diagnosing diseases and generating comprehensive radiology reports [\cite{jiang2024gpt}]. Taken together, these findings demonstrate that LLMs are not yet ready to serve as autonomous clinical decision-makers. While there may be potential for this in the future, the time is certainly not now.

\section*{Conclusion}

In our investigation, we compared GPT-4 with clinalytix Medical AI in predicting the clinical risk of developing delirium. Our results demonstrated that GPT-4 missed a significant number of positive cases compared to clinalytix Medical AI and was unable to reliably provide a probability for delirium risk. To understand the reasons behind these discrepancies, we further analyzed the LLM's outputs and identified potential explanations. These issues are not unique to our study or the specific model used; similar limitations have been observed in recent research, underscoring the difficulties LLMs face in accurately diagnosing conditions and interpreting complex clinical data. As a result, LLMs should not be relied upon as independent clinical decision-makers.

However, this does not diminish the potential of LLMs in healthcare. We believe they have substantial value and can offer significant support in the clinical domain. For now, LLMs should be used in assistive roles, augmenting the capabilities of clinicians and healthcare workers, rather than taking on fully autonomous decision-making responsibilities. Human supervision remains crucial to ensure optimal outcomes for patients and clinicians alike.

\bibliographystyle{apalike}
\bibliography{manuscript}

\end{document}